\documentclass{article}
\usepackage[english]{babel}

\usepackage[letterpaper,top=2cm,bottom=2cm,left=3cm,right=3cm,marginparwidth=1.75cm]{geometry}
\usepackage{CJK}

\usepackage{authblk}
\usepackage{amsmath}
\usepackage{graphicx}
\usepackage[colorlinks=true, allcolors=blue]{hyperref}
\date{Oct 1 2024} 

\title{Evaluating Deduplication Techniques for Economic Research Paper Titles with a Focus on Semantic Similarity using NLP and LLMs }

\author{Doohee You $^{1}$\footnote{Corresponding author: doohee@google.com},  Samuel P. Fraiberger $^{1, 2}$}

\affil{$^{1}$The World Bank\thanks{Disclaimer:This work utilizes data and/or methods potentially similar to those employed by the World Bank. The findings, interpretations, and conclusions expressed in this paper are entirely those of the author(s) and do not necessarily reflect the views of the World Bank or its affiliated entities.}, $^{2}$ New York University}

\begin{document}
\maketitle

\begin{abstract}
This study investigates efficient deduplication techniques for a large NLP dataset of economic research paper titles. We explore various pairing methods alongside established distance measures (Levenshtein distance, cosine similarity) and a sBERT model for semantic evaluation. Our findings suggest a potentially low prevalence of duplicates based on the observed semantic similarity across different methods. Further exploration with a human-annotated ground truth set is completed  for a more conclusive assessment. The result supports findings from the NLP, LLM based distance metrics.

In recent years, Large Language Models (LLMs) have garnered considerable attention for their remarkable abilities in natural language processing tasks. However, their widespread adoption has raised concerns pertaining to trust and safety. This systematic review investigates the current research landscape on trust and safety in LLMs, with a particular focus on the novel application of LLMs within the field of Trust and Safety itself. We delve into the complexities of utilizing LLMs in domains where maintaining trust and safety is paramount, offering a consolidated perspective on this emerging trend.

By synthesizing findings from various studies, we identify key challenges and potential solutions, aiming to benefit researchers and practitioners seeking to understand the nuanced interplay between LLMs and Trust and Safety. This review provides insights on best practices for using LLMs in Trust and Safety, and explores emerging risks such as prompt injection and jailbreak attacks. Ultimately, this study contributes to a deeper understanding of how LLMs can be effectively and responsibly utilized to enhance trust and safety in the digital realm.

\end{abstract}

\section{Introduction}

The World Bank is exploring the use of causal AI models for analyzing the impact of economic interventions. These models rely heavily on high-quality, accurate data, often derived from academic publications. However, large-scale NLP datasets can be susceptible to the presence of duplicate entries arising from: 1) Title variations: Authors may publish the same research with slightly different titles due to minor wording changes or phrasing variations. 2)Accidental duplicates: Duplicate submissions across academic databases can occur due to human error or technical glitches {Christen et al., 2012}\cite{christen2012csic}. 3)Near duplicates: Some research may be published with substantial content overlap but slightly different titles, requiring more sophisticated techniques for identification {Chevallier et al., 2022}\cite{chevallier2022detecting},{Shi et al., 2025}\cite{shi2025pretraineddatadeduplicationmodel}, and {Dhivyabharathi et al., 2016}\cite{7586397}.
Duplicate entries can negatively impact the functionality and accuracy of AI models. This study investigates various deduplication techniques using only titles as a starting point for building a clean and accurate dataset of RCT studies.

\section{Background}

Researchers have developed various techniques for deduplication in large NLP datasets, broadly categorized as follows: 1) String-based methods: These techniques rely on string similarity measures like Levenshtein distance {Miller, 2009}\cite{10.5555/1822502,} or Jaccard similarity to identify entries with high textual overlap {Gomaa et al., 2013}\cite{10.5120/11638-7118}. 2) Hashing methods: Hashing algorithms map similar textual data points to the same hash value, allowing for efficient duplicate detection. Locality-Sensitive Hashing (LSH) is a popular choice for large datasets {Dasgupta et al., 2011}\cite{dasgupta2011fast} 3) Embedding-based methods: These methods represent text data as numerical vectors using techniques like word embeddings or document embeddings. Similar entries will have vectors close in distance within the embedding space {Mikolov et al., 2013}\cite{mikolov2013}, {Di Liello et al., 2022}\cite{diliello2022pretrainingtransformermodelssentencelevel}.

\section{Method}

We evaluated various pairing techniques for comparing titles:
\begin{itemize}
\item{Complete Pairing Formula}

This method calculates the similarity between all possible pairs of titles within the dataset. The computational complexity is given by $\frac{n(n-1)}{2}$, where $n$ is the number of titles.

\item{Selective Pairing Across Sources}

This method focuses on comparing titles across different sources (e.g., JSTOR vs. ELSEVIER), excluding comparisons within the same source. This approach significantly reduces the number of comparisons.

\item{Pairing on Title Length}

This method considers titles with similar word counts to be more likely duplicates. We define a threshold ($\delta=5$) for the difference in word count between two titles {Aronovich et al., 2009}\cite{aronovich2009design}. Titles with word count difference less than or equal to $\delta$ are considered for further comparison.

Mathematically, for two titles $T_1$ and $T_2$ with word counts $w_1$ and $w_2$ respectively, this condition can be expressed as:

\begin{equation}
| w_1 - w_2 | \leq \delta
\end{equation}

\item{Pairing on Similar Length of Titles}

This method combines source information, word count ranges, and filtering based on the most common word length ($\mu$) in the dataset. Titles from different sources with word counts within a predefined range ($\lambda$) around $\mu$ are selected for comparison {Aronovich et al., 2009}\cite{aronovich2009design}. This condition mathematically as:

\begin{equation}
| T_1.word\_count - \mu | \leq \lambda \text{ AND } | T_2.word\_count - \mu | \leq \lambda
\end{equation}

\item{Pairing on Short Titles}

This approach focuses on titles with a length less than or equal to a predefined threshold ($\tau=3$), assuming simpler language and a higher chance of containing duplicates {Aronovich et al., 2009}\cite{aronovich2009design}. Mathematically, this can be expressed as:

\begin{equation}
T_1.word\_count \leq \tau \text{ AND } T_2.word\_count \leq \tau
\end{equation}

\end{itemize}

\subsection{Distance Measures}

Following title pre-processing and language detection, we utilized three distance measures:

\begin{itemize}
\item{Levenshtein Distance (LD)}

This metric calculates the minimum number of single-character edits (insertions, deletions, substitutions) needed to transform one string (title) into another {Miller, 2009}\cite{10.5555/1822502,}.

Formally, LD($T_1$, $T_2$) represents the Levenshtein distance between titles $T_1$ and $T_2$.

\item{Cosine Similarity (CS)}

This metric measures the similarity between two text vectors by calculating the cosine of the angle between them {Manning et al., 2008}\cite{manning2008}. A cosine similarity of 1 indicates identical vectors, while 0 represents orthogonal (completely dissimilar) vectors.

For two title vectors, $T_1$ and $T_2$, the cosine similarity is calculated as:

\begin{equation}
CS(T_1, T_2) = \frac{T_1 \cdot T_2}{||T_1|| ||T_2||}
\end{equation}

where:

* $\cdot$ represents the dot product between the two vectors
* $||T_1||$ and $||T_2||$ represent the magnitudes of the two vectors

\item{Pre-trained SBERT Model (Sentence Transformer)}

This method leverages a pre-trained model (SBERT) to generate semantic embeddings for the titles. The distance between the embeddings is used as a measure of similarity {Di Liello et al., 2022}\cite{diliello2022pretrainingtransformermodelssentencelevel}. SBERT models are trained on large datasets of text and sentence pairs, allowing them to capture semantic relationships between words and phrases.
\end{itemize}

\section{Results}

\begin{figure}[p]
  \centering
  \includegraphics[width=\textwidth]{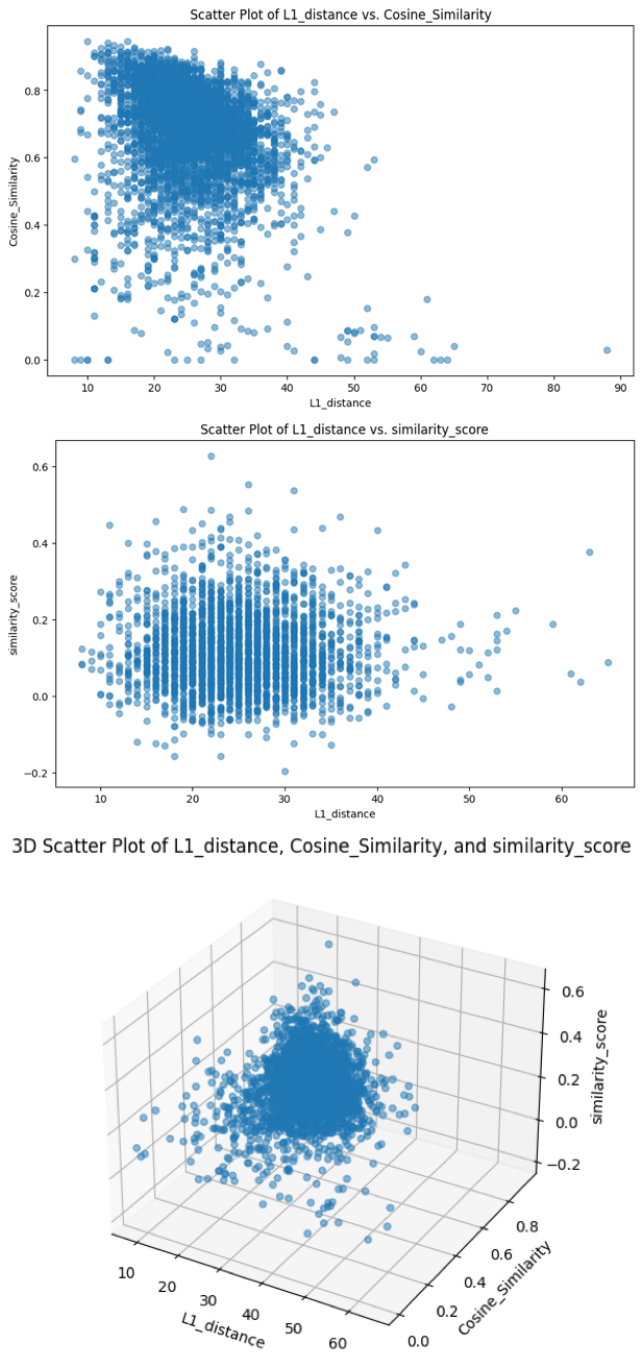}  
  \caption{Relationship between Distance Measurement Methods}
  \label{fig:distance_comparison}
\end{figure}

The relationship between three distance measurement methods (Levenshtein distance, cosine similarity, and SBERT model) is explored using 2,000 title pairs generated from a dataset of 11 million sources. The graph in Figure~\ref{fig:distance_comparison} visualizes how similar the distance scores are between these methods for the same title pairs.\\

The data points on the graph can exhibit different clustering patterns, revealing insights into the effectiveness of each method. Tight clustering suggests a strong correlation between the distance scores from all three methods, implying they often provide similar results in identifying similar or dissimilar titles. A scattered pattern, on the other hand, might indicate that the distance scores from different methods vary significantly for the same title pairs. This could suggest that each method has its own strengths and weaknesses in capturing different aspects of similarity. Additionally, method-specific clusters might emerge, providing further insights into their unique behavior. For instance, a cluster in the top left corner for Levenshtein distance could indicate many title pairs with very low edit distances, suggesting a high number of near-duplicates.\\

The ideal scenario for deduplication involves data points clustering in the bottom left corner of the plot. This would indicate that titles with low Levenshtein distance (very similar strings) also have low cosine similarity (semantically similar) and low BERT similarity scores (similar according to the BERT model). However, the graph in Figure~\ref{fig:distance_comparison} suggests a low prevalence of close matches using all three distance measures.\\

There is no strong cluster of points in the bottom left corner, indicating a lack of titles with very similar strings, low cosine similarity, and low BERT similarity scores. This suggests a low number of true duplicates within the dataset. Even titles with similar strings (low Levenshtein distance) might have varying semantics and score higher on cosine similarity or BERT similarity score. The scattered distribution on the Z-axis (BERT similarity) highlights this point. Even for titles with low Levenshtein distance and high cosine similarity (potentially similar wording and direction), the BERT similarity scores show variation. This suggests the BERT model might be capturing semantic differences between these titles that the other two metrics miss.\\

Several factors could contribute to this observation. The chosen BERT model might not be ideal for capturing the semantic nuances of economic research paper titles. Additionally, the thresholds used to define similar BERT similarity scores might need adjustment. Titles that appear similar based on string comparison and directional similarity might have subtle semantic differences. Finally, the analysis is based on a small sample size (between 1,000,000 to 2,000 samples out of 11 million titles) and might not represent the entire dataset. A larger sample might reveal tighter clustering if more duplicates are present.\\

\section{Conclusion}

This study investigated efficient deduplication techniques for a large NLP dataset of economic research paper titles, aiming to build a clean and accurate dataset for training causal AI models in economics. We explored various pairing methods alongside established distance measures like Levenshtein distance, cosine similarity, and pre-trained SBERT models for semantic evaluation.

\begin{itemize}
\item Novelty and Contributions:
While existing research explores deduplication techniques for NLP datasets, our work focuses specifically on economic research paper titles, a domain with unique characteristics. Previous studies often lack a thorough evaluation of different distance measures, particularly regarding semantic similarity for economic titles. Our contribution lies in the following aspects:
\item Comprehensive Evaluation of Distance Measures: We employed a diverse set of pairing techniques to identify potential duplicates and evaluated them using three established distance measures. This comprehensive approach provides valuable insights into the effectiveness of each measure for economic research paper titles.
\item Focus on Semantic Similarity: We leveraged a pre-trained SBERT model to capture semantic nuances beyond string similarity or directional similarity measured by Levenshtein distance and cosine similarity, respectively. This is particularly important for economic titles where subtle wording variations might not necessarily indicate duplicate content.
\item Ground-Truth Informed Analysis: To strengthen our evaluation, we created a ground-truth dataset of duplicates and non-duplicates. A subset of the selected title pairs were meticulously annotated by human raters on a one-by-one basis. This ensured the accuracy of our findings and allowed us to compare the results from all distance measures against a reliable benchmark.
\item Ethical Impact: While deduplication aims to improve data quality for AI models, it's crucial to consider potential ethical implications. Biases within the training data for deduplication models could lead to the removal of legitimate research, particularly from underrepresented areas of economic research. Future work should explore methods to mitigate bias and ensure fair treatment of diverse economic research titles.
\item Limitations:
Our analysis revealed a potentially low prevalence of close duplicates based on the observed semantic matching across different methods, which aligns well with the ground-truth data. However, developing domain-specific techniques tailored to economic research terminology might further enhance deduplication accuracy.
This research lays the groundwork for future exploration of advanced deduplication techniques specifically designed for economic research paper titles. By continuing to refine these methods, we can build high-quality datasets crucial for training robust causal AI models at the World Bank and across various economic research applications.
\end{itemize}

\bibliographystyle{alpha}
\bibliography{main}

\newcommand{\etalchar}[1]{$^{#1}$}
\begin{thebibliography}{MCCD13}

\bibitem[AAB{\etalchar{+}}09]{aronovich2009design}
Lior Aronovich, Ron Asher, Eitan Bachmat, Haim Bitner, Michael Hirsch, and Shmuel~T Klein.
\newblock The design of a similarity based deduplication system.
\newblock In {\em Proceedings of SYSTOR 2009: The Israeli Experimental Systems Conference}, pages 1--14, 2009.

\bibitem[Chr12]{christen2012csic}
Peter Christen.
\newblock Deduplication and data linkage in computer science.
\newblock {\em CSIC Technical Report TR-12-019}, 2012.

\bibitem[CRB{\etalchar{+}}22]{chevallier2022detecting}
Marc Chevallier, Nicoleta Rogovschi, Faouzi Boufar{\`e}s, Nistor Grozavu, and Charly Clairmont.
\newblock Detecting near duplicate dataset with machine learning.
\newblock {\em International Journal of Computer Information Systems and Industrial Management Applications}, 14:374--385, 2022.

\bibitem[DKS11]{dasgupta2011fast}
Anirban Dasgupta, Ravi Kumar, and Tam{\'a}s Sarl{\'o}s.
\newblock Fast locality-sensitive hashing.
\newblock In {\em Proceedings of the 17th ACM SIGKDD international conference on Knowledge discovery and data mining}, pages 1073--1081, 2011.

\bibitem[LGSM22]{diliello2022pretrainingtransformermodelssentencelevel}
Luca~Di Liello, Siddhant Garg, Luca Soldaini, and Alessandro Moschitti.
\newblock Pre-training transformer models with sentence-level objectives for answer sentence selection, 2022.

\bibitem[MCCD13]{mikolov2013}
Tomas Mikolov, Kai Chen, Greg Corrado, and Jeffrey Dean.
\newblock Distributed representations of words and phrases and their compositionality.
\newblock {\em Advances in neural information processing systems}, pages 1601--1608, 2013.

\bibitem[MRS08]{manning2008}
Christopher~D Manning, Prabhakar Raghavan, and Hinrich Schütze.
\newblock Introduction to information retrieval, 2008.

\bibitem[MVM09]{10.5555/1822502}
Frederic~P. Miller, Agnes~F. Vandome, and John McBrewster.
\newblock {\em Levenshtein Distance: Information theory, Computer science, String (computer science), String metric, Damerau?Levenshtein distance, Spell checker, Hamming distance}.
\newblock Alpha Press, 2009.

\bibitem[SLL{\etalchar{+}}25]{shi2025pretraineddatadeduplicationmodel}
Haochen Shi, Xinyao Liu, Fengmao Lv, Hongtao Xue, Jie Hu, Shengdong Du, and Tianrui Li.
\newblock A pre-trained data deduplication model based on active learning, 2025.

\bibitem[VK16]{7586397}
Dhivyabharathi~G V and S.~Kumaresan.
\newblock A survey on duplicate record detection in real world data.
\newblock In {\em 2016 3rd International Conference on Advanced Computing and Communication Systems (ICACCS)}, volume~01, pages 1--5, 2016.

\bibitem[WHG13]{10.5120/11638-7118}
Aly A.~Fahmy Wael H.~Gomaa.
\newblock A survey of text similarity approaches.
\newblock {\em International Journal of Computer Applications}, 68(13):13--18, April 2013.

\end{thebibliography}

\end{document}